\documentclass[10pt]{article} 
\PassOptionsToPackage{numbers, compress}{natbib}

\usepackage[preprint]{tmlr}

\usepackage[utf8]{inputenc} 
\usepackage[T1]{fontenc}    
\usepackage{hyperref}       
\usepackage{url}            
\usepackage{booktabs}       
\usepackage{amsfonts}       
\usepackage{nicefrac}       
\usepackage{microtype}      
\usepackage{xcolor}         
\usepackage[inline]{enumitem}
\usepackage{multirow}
\usepackage{subcaption}
\usepackage{pifont}

\usepackage{amsmath,amsfonts,bm}
\usepackage{xfrac}
\usepackage{cleveref}
\usepackage{mathtools}

\newcommand{\indicator}[1]{\mathbb{I} \Bigl(#1\Bigr)}

\def\eqref#1{equation~\ref{#1}}

\def\1{\bm{1}}

\DeclareMathAlphabet{\mathsfit}{\encodingdefault}{\sfdefault}{m}{sl}
\SetMathAlphabet{\mathsfit}{bold}{\encodingdefault}{\sfdefault}{bx}{n}

\DeclareMathOperator*{\argmin}{arg\,min}

\newcommand{\bbE}{\mathbb{E}}

\newcommand{\bbR}{\mathbb{R}}

\newcommand{\cL}{\mathcal{L}}

\newcommand{\cN}{\mathcal{N}}

\newcommand{\cT}{\mathcal{T}}

\newcommand{\cX}{\mathcal{X}}

\newtheorem{assumption}{Assumption}
\Crefname{assumption}{Assumption}{Assumptions}

\title{Supervised Learning Objectives for High-Dimensional\\Treatment Effects}

\author{%
  \name Nikita Dhawan \email nikita@cs.toronto.edu \\
  \addr University of Toronto, Vector Institute
  \AND
  \name Arnav Paruthi \email arnav.paruthi@mail.utoronto.ca \\
  \addr University of Toronto
  \AND
  \name Andrew Kim \email andrewkh.kim@mail.utoronto.ca \\
  \addr University of Toronto, Vector Institute
  \AND
  \name Lovedeep Gondara \email lovedeep\_gondara@vanguard.com \\
  \addr Vanguard
  \AND
  \name Jekaterina Novikova \email jekaterina\_novikova@vanguard.com \\
  \addr Vanguard
  \AND
  \name Chris J. Maddison \email cmaddis@cs.toronto.edu \\
  \addr University of Toronto, Vector Institute
}

\begin{document}

\maketitle

\begin{abstract}
Predicting the effect of interventions with many possible variations, \emph{e.g.}, therapeutic content that affects mental health outcomes or an earnings call transcript that drives movement in share price, is useful across several domains. However, classical causal estimators tend to assume that all possible interventions are observed, which is infeasible when interventions vary widely, for instance, in the space of all text strings. We adapt a well-known approach of recasting causal inference as a learning problem to address such settings, assuming realizability of the learned function and no unobserved confounding. We show that causal error decomposes into moment-balancing errors of increasing order and exploit structure of the conditional outcome in the confounders to design objectives that directly improve causal estimation. We also show how to project the effect of a high-dimensional treatment onto coarser attributes, which allows a single model to answer several causal questions without additional attribute-specific training. We empirically evaluate our estimators in settings with high-dimensional continuous, discrete, and text treatments. Our experiments demonstrate generalization to held-out treatments, benefits of balancing higher-order moments when required, and projected causal estimates that are competitive with attribute-specific ones.
Code is available at \href{https://github.com/nikitadhawan/causal-risk-minimization}{https://github.com/nikitadhawan/causal-risk-minimization}.
\end{abstract}

\section{Introduction}

Effective decision-making in real-world high-stakes domains often relies on predicting the effects of a combination of many different choices. 
For instance, the success of a therapy session may depend on the strategies employed, the order in which information is presented, the style of language used, and how these interact.  
Understanding the effect of all of these elements on a single outcome, \emph{e.g.}, patient well-being, is a causal estimation problem with a large space of possible interventions (or treatments), which in this case is the space of all strings.
Examples include a counselor's response to a patient in distress \citep{althoff2016large,ewbank2020quantifying}, the phrasing of a public health message \citep{milkman2021megastudy,dai2021behavioural}, word choices in advertisements \citep{pryzant2018deconfounded}, and the framing of an earnings disclosure \citep{hassan2020firm}. 

In this work, we focus on estimating the \emph{average potential outcome (APO)} of a high-dimensional treatment, which is defined as the average outcome that would be obtained under that treatment in a given population in the Neyman–Rubin model \citep{holland1986statistics, hirano2003efficient}.
Text strings as treatments is the key motivating setting, since understanding the effect of a whole text intervention can enable many types of decision-making: selecting optimal wording combinations, ranking candidate messages, or identifying linguistic strategies to achieve an outcome. 
The text treatment setting is especially challenging because treatments are very high-dimensional, each observed treatment may only occur once in the observed data, and treatments queried at test time may never have appeared during training. Even under standard causal assumptions that guarantee identifiability, estimation cannot rely on repeated observations of the same treatment and must share information across related treatments to generalize to unseen ones.

The traditional causal estimation approaches struggle when faced with high-dimensional treatments. Randomized controlled trials are too expensive or infeasible to conduct while covering the entire space of possible treatments. 
Na\"ive applications of classical observational estimators to this setting also face several challenges. Inverse propensity score weighting (IPW) \citep{horvitz1952generalization,rosenbaum1987model} computes an empirical sum that is restricted to units that received the exact treatment of interest, and is undefined for any string not observed in the training data. Outcome imputation \citep{rubin1977assignment} can sidestep this by learning a model of the conditional outcome given treatment and confounders and marginalizing over the confounders. However, this requires inference-time access to individual-level confounder values, which may be impractical when those variables are sensitive (\emph{e.g.}, income or health markers) and prohibitively expensive when the confounders themselves are high-dimensional (\emph{e.g.}, textual confounders). 
Adaptations of these methods rely on additional structure, such as a parametric model of treatment assignment \citep{hirano2005propensity} or sparsity \citep{kramer2026causal}, to handle high-dimensional settings.

The approach we adapt is to recast causal estimation as a learning problem so that a model trained across observed texts can generalize APO estimates to held-out treatments, under the assumption that the APO function is realizable in a model class over treatments.
It is known that causal estimation can be recast as weighted risk minimization \citep{swaminathan2015counterfactual,hassanpour2019counterfactual,jung2020learning}, a general approach we call \emph{Causal Risk Minimization (CRM)} throughout, but works in the CRM literature tend to focus on binary treatments, which obscures some of the challenges with high-dimensional ones.

\begin{figure}[t]
    \centering
    \includegraphics[width=\textwidth]{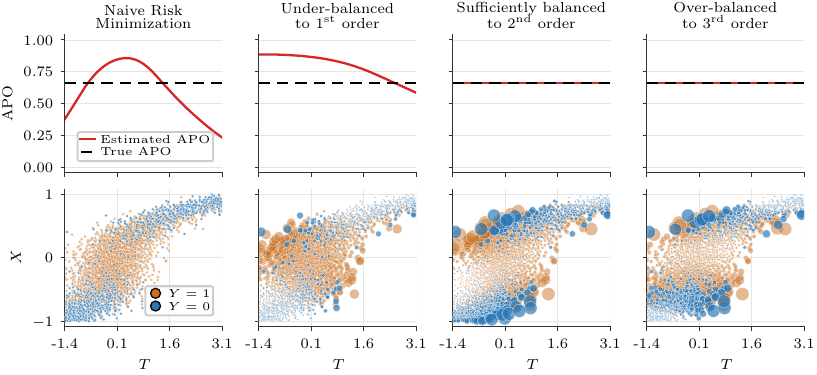}
    \caption{In this toy example, the observed outcome $Y$ is quadratic in confounder $X$, which is in turn correlated with treatment $T$. Hence, even though the true APO is constant in $T$, standard risk minimization results in a biased APO estimator. Weighted risk minimization with weights, visualized by marker sizes, improves estimation as they balance moments up to the second order. Balancing the third moment as well changes the weights but not APO estimation, and hence is unnecessary.}
    \label{fig1_balance}
\end{figure}

Our key contribution is to address two of the challenges faced by causal risk minimization when applied to high-dimensional treatments.
First, APO estimation via risk minimization requires targets that must themselves be estimated from data. In the case of IPW, the targets are weighted outcomes, such that APO estimation quality depends directly on the correctness of the learned weights.
Empirical or trained models of these weights often run the risk of producing extreme weights which in turn harm APO training and prediction. 
We address this by showing that when the conditional outcomes vary smoothly enough as a function of the confounders, the APO estimation error decomposes into a series of \emph{moment-balancing errors} — errors in balancing the moments of different orders of the confounder distribution across treatment groups. This decomposition, derived for binary, discrete, and continuous confounders, motivates learning importance weights with explicit balance regularization up to an order that depends on the smoothness of the outcome in confounders. While prior works have either used low-order balance regularization to improve propensity score training \citep{imai2014covariate} or attempted full distributional balancing \citep{huling2024independence}, our error decomposition provides a principled strategy to choose the order up to which balancing confounder moments improves APO estimation itself.
For instance, in the toy demonstration of \cref{fig1_balance}, balancing moments upto the second order consistently improves APO estimation. Balancing the third-moment results in different weights that balance the full confounder distribution better but does not improve APO estimation any further.

Second, practical use-cases of an APO estimator may include asking lower-dimensional causal questions about the effects of specific text attributes on an outcome. 
Prior work using texts as treatments has largely circumvented high-dimensionality issues by collapsing the text into a low-dimensional, often binary, summary before classical estimation \citep{fong2016discovery,pryzant2018deconfounded,pryzant2020causal,zhang2020quantifying}. However, this approach may sacrifice valuable information, requires committing to the textual attributes of interest \emph{a priori}, and involves training a separate estimator per attribute. Instead, we demonstrate how to retain the full (text) treatment space and efficiently project a single high-dimensional APO estimator onto lower-dimensional attributes \emph{post hoc}, given a model of the relationship between the attributes and text. This allows us to simultaneously answer several lower-dimensional causal questions, at no additional training cost per attribute.

We evaluated our CRM estimators across three settings of increasing complexity: a simple linear setting with continuous treatments, a synthetic dataset with high-dimensional discrete treatments, and a semi-synthetic dataset of real Amazon Reviews \citep{hou2024bridging} with 10,000 unique text treatments. 
For text treatments, we finetuned a pretrained language model on the CRM loss function that is aligned with common pretraining objectives and hence, can harness the priors that these models encode over natural language \citep{choi2210lmpriors,capstick2411autoelicit}. 
Our experiments empirically verify our structural assumptions and the analytic APO error decomposition, demonstrate generalization to held-out treatments, and tease apart the effect of different experimental parameters on prediction quality. 
In particular, optimizing higher-order moment-balancing errors according to our error decomposition consistently improved our weighted CRM estimator.
In the text setting, where every review is observed only once, CRM estimators nevertheless generalized APO predictions to held-out reviews and approached oracle performance, but without access to confounders at inference-time.
Finally, projections of our Amazon Reviews APO estimator onto attributes like sentiment and review length matched or outperformed bespoke estimators that were trained separately per attribute. 
These results suggest that causal estimation over the full space of text treatments can be feasible even when individual treatments do not repeat, provided that outcomes generalize across treatments through the chosen model class and depend sufficiently smoothly on confounders so that a tractable order of moment balance suffices.

\section{Preliminaries}
\label{sec:background}

Let $\cT$ be the space of possible treatments and let $Y(t)$ for $t \in \cT$, $X$, and $T$ be random variables corresponding to the potential outcomes, the confounders, and the treatment, respectively, all drawn from some joint distribution. We assume $Y(t)$ takes values in either $\bbR$ or $\{0, 1\}$ and $X$ takes values in $\cX$. Our goal is to estimate the \emph{average potential outcome} (APO) under treatment $t$,
\begin{equation}
    g(t) = \bbE[Y(t)],
\end{equation}
as defined in the Neyman–Rubin model \citep{holland1986statistics}.
We assume access to $n$ i.i.d. samples $(Y_i, X_i, T_i)$ from an observational distribution, where we abuse notation and let $Y = Y(T)$ be the random potential outcome realized under the observed treatment $T$. 
Since the treatment assignment may depend on $X$, the na\"ive conditional mean $\bbE[Y \mid T = t]$ is in general biased for $g(t)$, as illustrated in \cref{fig1_balance}. Throughout, we assume strong ignorability (\emph{i.e.} no unmeasured confounding):
$\{Y(t)\}_{t \in \cT} \perp T \mid X$,
positivity: $0 < p_{T \mid X}( t \mid x ) < 1$,
and the stable unit treatment value assumption \citep{rubin1980randomization},
under which the following classical estimators are valid.
We provide a complete list of notation in \cref{app:notation}.

\textbf{Inverse Propensity Score Weighting (IPW).}
The IPW estimator \citep{horvitz1952generalization,rosenbaum1987model} uses an estimate $\hat{e}(t, x)$ of the \emph{propensity score} $p_{T \mid X}(t \mid x)$ to estimate APOs as:
\begin{equation}
\label{eq:ipw_classic}
    \hat{g}_{\text{IPW}}(t) = \frac{1}{n} \sum_{i=1}^n \frac{Y_i \, \indicator{T_i = t}}{\hat{e}(T_i, X_i)}.
\end{equation}
Notice that the indicator $\mathbb{I}(T_i = t)$ restricts the sum above to units that received treatment $t$. 
In high-dimensional settings like those with text treatments, we may only ever observe one datapoint per treatment, leading to stability issues. Even when positivity is satisfied, there may be test-time treatments that were not seen in the training data, in which case this estimator is undefined.

\textbf{Outcome Imputation (OI).}
The outcome imputation estimator \citep{rubin1977assignment} learns an estimate $\hat{f}(t, x)$ of the expected conditional outcome $\bbE[Y \mid T=t, X=x]$ to estimate APOs as: 
\begin{equation}
\label{eq:oi_classic}
    \hat{g}_{\text{OI}}(t) = \frac{1}{n} \sum_{i=1}^n \hat{f}(t, X_i).
\end{equation}
While $\hat{f}$ may generalize to held-out $t$ with a correctly specified model class, the Monte Carlo average over $n$ confounder values for each treatment $t$ can be computationally expensive at inference time. Moreover, when confounders $X$ are sensitive (\emph{e.g.}, income or health status), access to individual-level $X_i$ at inference time may not be feasible for privacy reasons.

While the standard causal assumptions above guarantee that $g$ is identifiable, practical settings make estimation challenging or even impossible without additional structural assumptions, such as parametric forms \citep{hirano2005propensity}, sparsity \citep{kramer2026causal}, or smoothness of $g$ that enables sharing information across treatments when not all of them are observed during training.
These observations motivate the development of estimators that
\begin{enumerate*}[label=(\roman*)]
    \item share information across treatments by exploiting the structure in $\cT$ and the inductive bias of trained models, and 
    \item produce APO estimates $\hat{g}(t)$ as a direct function of $t$ alone.
\end{enumerate*}

\section{Causal Risk Minimization}
\label{sec:method}

Many causal estimation tasks can be framed as risk minimization \citep{swaminathan2015counterfactual,hassanpour2019counterfactual,jung2020learning}, an approach we collectively call \emph{Causal Risk Minimization} (CRM). \Cref{fig1_balance} demonstrates with a toy example how weighted risk minimization recovers true APOs given sufficiently balancing weights. The core challenge in CRM is learning $g$ from finite data; prior work typically imposes regularity conditions to ensure that minimizing empirical risk actually recovers the true population minimizer (in the limit). We broadly follow the same philosophy, but instead of committing to a specific regularity condition, we note that this kind of finite-sample generalization is widely studied in machine learning and capture it informally in the following assumption:

\begin{assumption}
    \label{assump:realizability}
    (Informal)
    Treatments, observed and held-out, are drawn from a marginal distribution $p_T$ on $\cT$ and the APO function $g(t)$ for $t \in \cT$ belongs to a known hypothesis class that is regular enough to enable generalization from finite samples.
\end{assumption}
In the case of text treatments, we will use a fine-tuned language model to realize this class.
Because this learning algorithm is too complex to provide precise assumptions for, we do not formally characterize finite-sample estimation or optimization errors, and instead study it empirically in \cref{sec:experiments}.

The critical quantity in this estimation problem is a weight function, whose estimation is the focus of one of our contributions. Our second key contribution is a simple Monte Carlo method for computing APO functions of treatments projected to lower dimensions.
For text treatments in particular, the cross-entropy loss variant derived below allows us to finetune pretrained language models on an objective that aligns with their next-token-prediction pretraining objective. 

\subsection{Deriving CRM Targets}
\label{crm_targets}
\textbf{IPW-CRM.}
The IPW estimator in \cref{eq:ipw_classic} is limited by the requirement that treatment $t$ appear in the observed data. Note that, under strong ignorability and with true propensity scores,
\begin{equation}
\label{eq:ipw_crm}
\begin{aligned}
g(t) 
&= \bbE\left[\frac{Y \indicator{T = t}}{p_{T \mid X}(T \mid X)} \right] 
= \bbE\left[\bbE\left[\frac{Y \indicator{T = t}}{p_{T \mid X}(T \mid X)} \, \middle| \, T \right]\right] \\
&= \int p_T(s) \bbE\left[\frac{Y \indicator{T = t}}{p_{T \mid X}(T \mid X)} \, \middle| \, T=s \right] ds 
= \bbE\left[\frac{Y p_T(t)}{p_{T \mid X}(t \mid X)} \, \middle| \, T=t \right],
\end{aligned}
\end{equation}
where the first three equalities follow, respectively, from the standard IPW estimator, the law of total expectation, and the definition of expectation with respect to $T$. The final equality holds because in the integral over $s \in \mathcal{T}$, the indicator variable $\mathbb{I} (T=t)$ evaluates to $1$ when $s=t$ and $0$ otherwise.   
\Cref{eq:ipw_crm} motivates fitting a parametric IPW-CRM model $\hat{g}(t)$ with the following objective:
\begin{equation}
\label{eq:ipw_loss}
    \hat{g}_{\text{IPW-CRM}} = \argmin_g \sum_{i=1}^n \cL \left(g(T_i), \frac{\hat{p}_T(T_i)}{\hat{e}(T_i, X_i)} \, Y_i \right),
\end{equation}
where $\hat{e}(T_i, X_i)$ is a learned propensity score, $\hat{p}_T(T_i) = \frac{1}{n}\sum_{j=1}^n \hat{e}(T_i, X_j)$ is the marginal treatment probability estimate, and $\cL$ is a suitable loss function. For real $Y$, we can take $\cL$ as squared error, and for binary $Y$, we take it as cross-entropy loss with pseudo-targets $[1 - Y_i \cdot \frac{\hat{p}_T(T_i)}{\hat{e}(T_i, X_i)}, Y_i \cdot \frac{\hat{p}_T(T_i)}{\hat{e}(T_i, X_i)}]$.

\textbf{SW-CRM.}
The weighted-outcome targets in \cref{eq:ipw_loss} are computed by first learning a model for propensity scores. However, we may also directly learn a model of the stabilized weights $\hat{w}(T_i, X_i)$ \citep{robins2000marginal} to then fit a Stabilized-Weighted APO estimator (SW-CRM) given by:
\begin{equation}
\label{eq:sw_loss}
    \hat{g}_{\text{SW-CRM}} = \argmin_g \sum_{i=1}^n \cL \left(g(T_i), \hat{w}(T_i, X_i) \, Y_i \right),
\end{equation}
where the loss function $\cL$ is chosen similarly to that for IPW-CRM.

\textbf{OI-CRM.}
The OI estimator of \cref{eq:oi_classic} required fitting a conditional outcome model $\hat{f}$ and then estimating APOs with $\hat{g}(t) = \frac{1}{n}\sum_{i=1}^n \hat{f}(t, X_i)$, assuming access to individual-level confounder values and requiring multiple ($n$) forward passes during inference. Given a trained $\hat{f}$, we can lift this requirement by training an OI-CRM estimator: 
\begin{equation}
\label{eq:oi_loss}
    \hat{g}_{\text{OI-CRM}} = \argmin_g \sum_{i=1}^n \cL \left(g(T_i), \frac{1}{n}\sum_{j=1}^n \hat{f}(T_i, X_j) \right),
\end{equation}
where the loss function $\cL$ is squared error.
Note that all estimators learned via \cref{eq:ipw_loss,eq:sw_loss,eq:oi_loss} learn a direct mapping from treatments to APOs, do not assume access to confounder values at inference time, and require only a single forward pass to predict the APO for a given treatment.

\subsection{APO Error Decomposition}
\label{error_decomp}

Under \cref{assump:realizability} and in the infinite-data limit, the solutions of \cref{eq:ipw_loss,eq:sw_loss,eq:oi_loss} recover $g$, provided the targets used in the loss functions are correct. In practice, these are computed using a learned model of propensity scores, stabilized weights, or conditional outcomes, respectively. However, even small errors in calibration of these learned models, a common pitfall of neural networks, can lead to large errors in APO estimation. In the case of propensity scores and stabilized weights, we can decompose the error in APO estimation into a series of balancing error terms. Confounder balancing is a well-known necessary condition for correct propensity scores \citep{imai2014covariate}. The error decomposition below shows which balance conditions matter for APO estimation error via risk minimization, based on the structure of the conditional outcome in the confounders. We present the decomposition in terms of the stabilized weights $w(t, x)$, while the same analysis holds for propensity scores $e(t, x)$.

The balancing property of stabilized weights states that they balance any function $h$ of the confounders $X \in \cX$ across treatments: $\bbE [ w(t, X) \, h(X) | T=t] = \bbE[h(X)]$ for all $t \in \cT$. Perfect APO estimation via the optimization in \cref{eq:sw_loss} requires the learned $\hat{w}(t, X)$ to satisfy this for $h(X) = f(t, X)$.
\begin{equation}
\label{eq:f_balance}
\begin{aligned}
    \hat{g}_{\text{SW-CRM}}(t) &= \bbE[\hat{w}(T, X) \, Y \mid T = t] = \bbE[ \bbE[ \hat{w}(T, X) \cdot Y \mid T=t, X ] \mid T = t ] \\
    &= \bbE[ \hat{w}(t, X) \bbE[ Y \mid T=t, X ] \mid T = t ] = \bbE[ \hat{w}(t, X) \, f(t, X) \mid T = t ].\\
    \hat{g}_{\text{SW-CRM}}(t) &= g(t) = \bbE[f(t, X)] \iff \bbE[ \hat{w}(t, X) \, f(t, X) \mid T = t ] = \bbE[f(t, X)].
\end{aligned}
\end{equation}

Thus, APO estimation does not require balance of the confounder distribution for its own sake. It requires the learned weights to balance $f(t, X)$ for all $t \in \cT$. Since $f$ is unknown, existing methods must choose which surrogate functions of $X$ to balance \citep{peyrot2026choosing,huang2022higher}. A common choice is the first-moment balance \citep{imai2014covariate,athey2018approximate}, which is sufficient under a linear outcome model, while richer approaches balance broad function classes to approximate full distributional balance \citep{du2021adversarial,huling2024independence}.
Instead, we interpolate between these two extremes and make the following assumption on the structure of $f(t, \cdot)$, to relate APO estimation error to moment-balancing errors of increasing order.
Here, the structure of the outcome function in the confounders determines the smallest sufficient moment order, up to which minimizing balance errors directly minimizes APO error.

\begin{assumption}
\label{assump:poly}
For every $t \in \cT$, either (i) the conditional outcome $f(t, \cdot)$ is a polynomial of degree at most $K$, or (ii) the confounder $X$ has finite support $\{0, 1, \ldots, K\}$.
\end{assumption}
In either case, $f(t, X)$ admits an expansion in powers of $X$ up to order $K$. Hence, we can write $f(t,X) = \sum_{k=0}^K c_k(t)\,X^k)$ for constants $c_k(t)$, and \cref{eq:f_balance} gives:
\begin{equation}
  \hat{g}(t) - g(t) = \sum_{k=0}^K c_k(t) \cdot \varepsilon_k(t), \text{ where }  
  \varepsilon_k(t) = \mathbb{E} \left[\hat{w}(T,X) \cdot X^k \mid T=t\right] - \mathbb{E}[X^k],
  \label{eq:general-decomp}
\end{equation}
is the \emph{order-$k$ balancing error} corresponding to $X^k$. We derive this decomposition for binary, discrete, and continuous confounders in \cref{app:error_decomp}
This error decomposition suggests a training scheme for learning stabilized weights to minimize all balancing errors up to order $K$:  
\begin{equation}
    \hat{w} = \argmin_{w} \sum_{t \in \{T_1, \dots, T_n\}} \sum_{k=0}^{K} \left\| \frac{1}{n_t} \sum_{i: T_i = t} w(t, X_i) \cdot X_i^{k} - \frac{1}{n} \sum_{j=1}^{n} X_j^{k} \right\|^{2},
\end{equation}
where $n_t = |\{i : T_i = t\}|$ is the number of units that received treatment $t$.
For IPW-CRM, we add this as a regularization term to the standard propensity score training objective. 
For binary confounders $(X \in \{0,1\})$, for example, normalization and first-moment balancing ($K=1$) are sufficient.
Note that for smooth non-polynomial $f$, truncating the above decomposition at order $K$ results in a Taylor remainder controlled by the $(K+1)$-th derivative of $f$. Hence, smoother $f(t, \cdot)$ can make small $K$ sufficient even when no finite-order expansion is exact.

\subsection{Lower-dimensional projections of high-dimensional APO estimates}
\label{projection}
How do we use high-dimensional APO estimates in practice? In the case of text treatments, for instance, we may compare APOs of two texts or rank some number of text candidates to select the best one for an outcome of interest. We may also be interested in the effects of some lower dimensional attribute, or coarsening, of the text, such as length or sentiment, in order to deduce general strategies that favor an outcome. We denote this attribute as $T'$ and assume it satisfies the following.
\begin{assumption}
\label{assump:projection}
    A known conditional distribution $p_{T' \mid T}$ relates $T'$ and $T$: $p_{T'}(t') = \bbE_T [p_{T'|T}(t'|T)]$. Further, $T'$ is conditionally independent of $X$ and $Y(\cdot)$ given $T$: $T' \perp (X, Y(\cdot)) \mid T$.
\end{assumption}
This assumption holds trivially for deterministic properties of the treatment, for instance, mention of particular words in the text or the length of the text.
Under \cref{assump:projection}, the lower-dimensional APO $g'(t')$ is given by the expectation of high-dimensional APOs $g(\cdot)$ over samples where $T' = t'$, and can be computed via a Monte Carlo estimate as:
\begin{equation}
\label{eq:proj}
    g'(t') = \bbE_{T| T'=t'} [g(T)] \approx \frac{\sum_{i=1}^n \indicator{T'_i=t'} \, \hat{g}(T_i)}{\sum_{i=1}^n \indicator{T'_i=t'}}.
\end{equation}
See \cref{app:projection} for a complete description of the lower-dimensional projections of treatments and derivation of their APOs.
\Cref{eq:proj} enables the use of a single learned $\hat{g}$ to simultaneously answers causal questions about any lower-dimensional attribute that can be labeled given the high-dimensional treatment, without retraining a separate estimator per attribute.
\section{Related Work}

\textbf{High-dimensional treatments.}
Causal inference with complex, multi-valued, or high-dimensional treatments is a well-recognized challenge \citep{lopez2017estimation,wang2026causal}. \citet{hirano2005propensity} defined the generalized propensity score (GPS) to extend propensity-score-based techniques to continuous treatments, where no treatment value recurs and estimation requires smoothness of the dose-response function. 
\citet{kennedy2017non} developed doubly robust pseudo-outcome regression for a similar setting. \citet{kramer2026causal} proved that meaningful estimation with very high-dimensional treatments is impossible without sparsity assumptions, even when each treatment is observed more than once. Our approach instead assumes structure in the APO function itself (see \cref{assump:realizability}). For text treatments, pretrained language models provide a natural model class that, after finetuning with the right targets, can model semantic similarity of texts and generalize to held-out ones.

\textbf{Text treatments.} 
A growing literature applies these ideas when the treatment is expressed in natural language \citep{feder2022causal}, though prior work has largely circumvented the high-dimensionality of text by first extracting lower-dimensional representations \citep{imai2024causal}. These include lexical features \citep{pryzant2018deconfounded}, inferred latent variables \citet{fong2016discovery,pryzant2020causal}, bag-of-words representations \citep{sharma2020hi}, and underlying binary treatment variables \citep{guo2024estimating}.
\citet{zhang2020quantifying} similarly reduced counselor conversations to a handful of attributes (\emph{e.g.} length, speed, sentiment).  
\citet{wood2018challenges} highlighted the challenges of using text classifiers for such dimensionality reduction.
These approaches can be summarized as constructing a setting with lower-dimensional treatments and multiple observations of each, at the cost of committing to the dimensions of interest \emph{a priori} and discarding relevant treatment information. 
We instead retain the full space of strings and treat each unique text as a distinct treatment. Lower-dimensional causal questions can then be specified \emph{post hoc} and answered by projecting the resulting full-text APO estimator.
Large language models also increasingly appear in causal pipelines with text in other roles, for structuring patient self-reports \citep{dhawan2024end} or controlling for it as a confounder \citep{ma2025llm}.
Applications motivating the study of text treatments span finance \citep{zhou2023causal,yang2025estimating,ongsakul2025leveraging,hassan2020firm}, consumer behaviour \citep{yang2025estimating,pryzant2020causal} and mental health \citep{bertagnolli2020counsel,ewbank2020quantifying,althoff2016large}. 
Across all these settings, prior work has relied on manual annotation or classification to reduce treatments to lower dimensions \emph{a priori}.

\textbf{Covariate balancing.}
Correct propensity scores balance functions of confounders across treatments, a property that has inspired several methods to learn balancing weights for binary \citep{imai2014covariate,hainmueller2012entropy,wong2018kernel} or continuous \citep{fong2018covariate} treatments.
A central practical difficulty is deciding which functions of confounders to balance \citep{peyrot2026choosing}. Existing methods target either low-order moments \citep{athey2018approximate} or all possible functions to achieve full independence between confounders and treatments \citep{ai2021unified,huling2024independence}. Kernel approaches \citep{hartman2025kpop} argue in favour of the latter when little is known about outcome structure. However, as shown in \cref{error_decomp}, this can be unnecessary when we have information about the relevant dependence on confounders.
\citet{kallus2020generalized} developed a generalized optimal matching framework that subsumes covariate balance and establishes a balance-variance trade-off. 
Other approaches include adversarial balancing \citep{du2021adversarial,kallus2020deepmatch,ozery2018adversarial,kazemi2024adversarially}, end-to-end stable weight learning \citep{bahadori2022end}, and joint propensity-outcome training with targeted regularization \citep{shi2019adapting}.
Our decomposition exploits \cref{assump:poly} and relates APO estimation error directly to moment-balancing errors up to an order determined by the smoothness of the outcome model in confounders, rather than being a free parameter of the method.
Further, while the majority of the works above were evaluated on either binary or continuous treatments, our estimators apply to high-dimensional discrete treatments, including text.

\section{Empirical Evaluation}
\label{sec:experiments}

Our experiments investigated the following questions:
\begin{enumerate*}[label=(\roman*)]
\item Does balance regularization improve direct APO estimation in a simple linear setting with continuous treatments?
\item How do the variants of our method perform in synthetic settings with high-dimensional discrete treatments, with varying sample size, model size, confounding strength, and order of moments balanced?
\item Do our methods scale to more realistic settings to predict potential outcomes of text treatments using large language models?
\item Can we repurpose a single high-dimensional APO estimator to answer several lower-dimensional causal questions, without training separate estimators for each?
\end{enumerate*}

\textbf{Datasets.}
We evaluated different methods on three settings of increasing complexity.
First, we verified our error decomposition in a simple linear setting with continuous Gaussian confounders, treatments, and outcomes, where the true APO is a linear function of confounders and the treatment.
Next, we constructed a synthetic setting with discrete confounders, binary outcome, high-dimensional discrete treatments to mirror natural language tokens, and nonlinear relationships between different variables.
Finally, we evaluated different methods on a real-world dataset of Amazon Reviews \citep{hou2024bridging} in a semi-synthetic setting, where each product review is a different treatment, total number of ratings received by the product is the $8$-dimensional confounder, and the purchase of the product by a user reading a given review is the binary outcome of interest. Outcomes were synthetically generated by using GPT-5.1 \citep{singh2025openai} as a conditional outcome model, which we prompted for a realistic probability of purchase, given each possible confounder value and treatment review. All datasets were split into training, validation, and test subsets such that treatments are non-overlapping across the subsets. In the case of Amazon Reviews, each treatment (or review) is observed only once.
Complete details for all datasets and experimental parameters are provided in \cref{app:dataset_details,app:exp_details}, respectively.

\textbf{Evaluation.}
Across all settings we evaluated APO estimation quality via the relative mean absolute error (MAE) and correlation between estimated and true APOs over held-out treatments. For estimators that measure and optimize balance errors, we also report the average of balance errors up to some order $K$, where $K$ is specific to each setting. 

\begin{figure}[t]
    \centering
    \includegraphics[width=\linewidth]{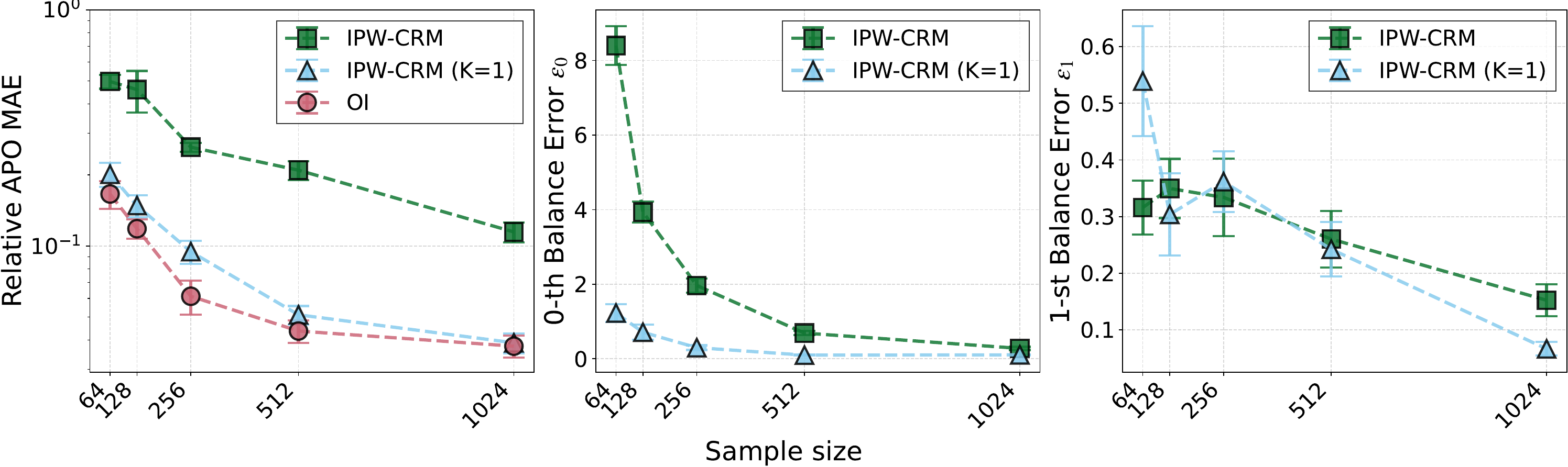}
    \caption{Balancing the order-0 and order-1 moments significantly improved IPW-CRM performance, matching that of the oracle OI, for linear models and continuous treatments and covariates.}
    \label{fig:gaussian}
\end{figure}

\subsection{Does optimizing balance errors reduce APO error in a linear continuous setting?}
\label{sec:gaussian}
We used continuous Gaussian data with a linear true APO function to empirically verify the APO error decomposition into the order-0 and order-1 balance errors, as analytically derived in \cref{sec:gaussian_details}.
Notice that the continuous setting requires generalization to held-out treatments, as in the case of text treatments.
We compared IPW-CRM estimators with and without balance regularization against an oracle outcome imputation estimator (OI) that trains a correctly specified linear network to model the conditional outcomes. 
\Cref{fig:gaussian} shows relative APO MAE, alongside the corresponding order-0 and order-1 balance errors for the IPW-CRM estimators, all of which reduce as sample size increases for all estimators. Without any balance error regularization, IPW-CRM suffered larger balance errors and corresponding APO estimation errors. In contrast, the balance error regularizer used in IPW-CRM with $K=1$ substantially reduced balancing errors, which in turn reduced APO error, as expected, and matched the oracle performance of OI without its inference-time access to $X$. This empirically verifies that minimizing moment-balancing errors directly translates to reduced APO error.

\newcommand{\cmark}{\ding{51}}%
\newcommand{\xmark}{\ding{55}}%

\begin{table}[t]
    \centering
    \scriptsize
    \caption{On both, synthetic and Amazon Reviews, datasets, SW-CRM outperformed other estimators, with higher-order moment balancing improving performance across metrics.}
    \resizebox{\linewidth}{!}{
    \begin{tabular}{lcc cc cc}
    \toprule
    & &  & \multicolumn{2}{c}{\textbf{Synthetic Discrete}} & \multicolumn{2}{c}{\textbf{Amazon Reviews}} \\
    \cmidrule(lr){4-5} \cmidrule(lr){6-7}
    \textbf{Estimator} & $K$ & \textbf{ Test-time $X$} & \textbf{Rel.\ MAE} & \textbf{APO Correlation} & \textbf{Rel.\ MAE} & \textbf{APO Correlation} \\
    \midrule
    OI (Oracle)  & $-$ & \cmark & $0.184$ & $0.943$ & $0.167$ & $0.830$ \\
    \midrule
    OI-CRM        & $-$ & \xmark & $0.143$ & $0.969$ & $0.226$ & $0.812$ \\
    IPW-CRM                 & $1$ & \xmark & $0.166$ & $0.901$ & $0.871$ & $0.190$ \\
    SW-CRM        & $1$ & \xmark & $0.141$ & $0.935$ & $0.279$ & $0.824$ \\
    SW-CRM        & $2$ & \xmark & $\mathbf{0.121}$ & $\mathbf{0.996}$ & $\mathbf{0.220}$ & $\mathbf{0.874}$ \\
    \bottomrule
    \end{tabular}
    }
    \label{tab:combined}
\end{table}

\subsection{How does CRM perform with discrete treatments and varying experimental parameters?}
\label{sec:discrete}
As a step towards high-dimensional text treatments, we used a synthetic dataset of high-dimensional discrete treatments and trained small transformer models to compare the performance of different estimators. As shown in \cref{tab:combined}, SW-CRM ($K=2$) outperforms vanilla OI and other variants. Including higher-order moment-balancing regularization improves its performance. For IPW-CRM, we report results for the best performing $K=1$, which is equivalent to standard low-order moment balancing methods \citep{imai2014covariate}. For larger $K$, IPW-CRM suffered from training instability and hence, poorer performance, a pitfall avoided by SW-CRM. 

\begin{figure}[t]
\centering
\scriptsize
  \begin{subfigure}[t]{.32\textwidth}
    \centering
    \includegraphics[width=\linewidth]{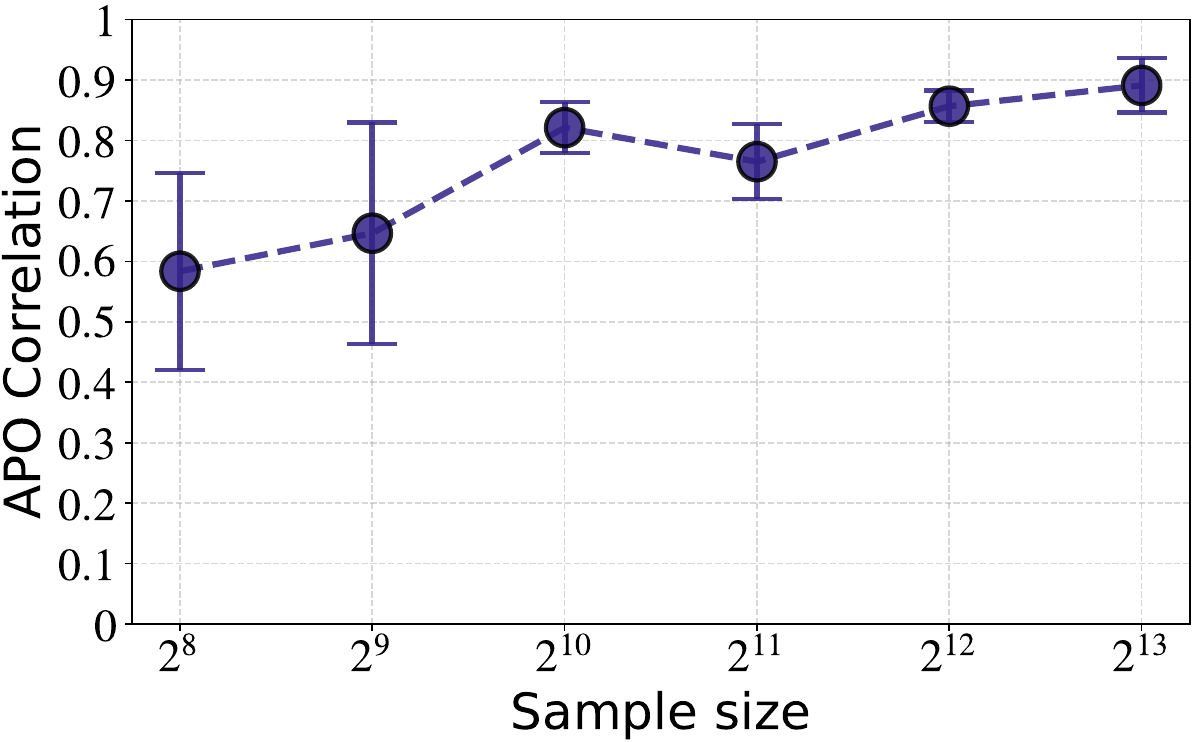}
  \end{subfigure}
  \hfill
  \begin{subfigure}[t]{.32\textwidth}
    \centering
    \includegraphics[width=\linewidth]{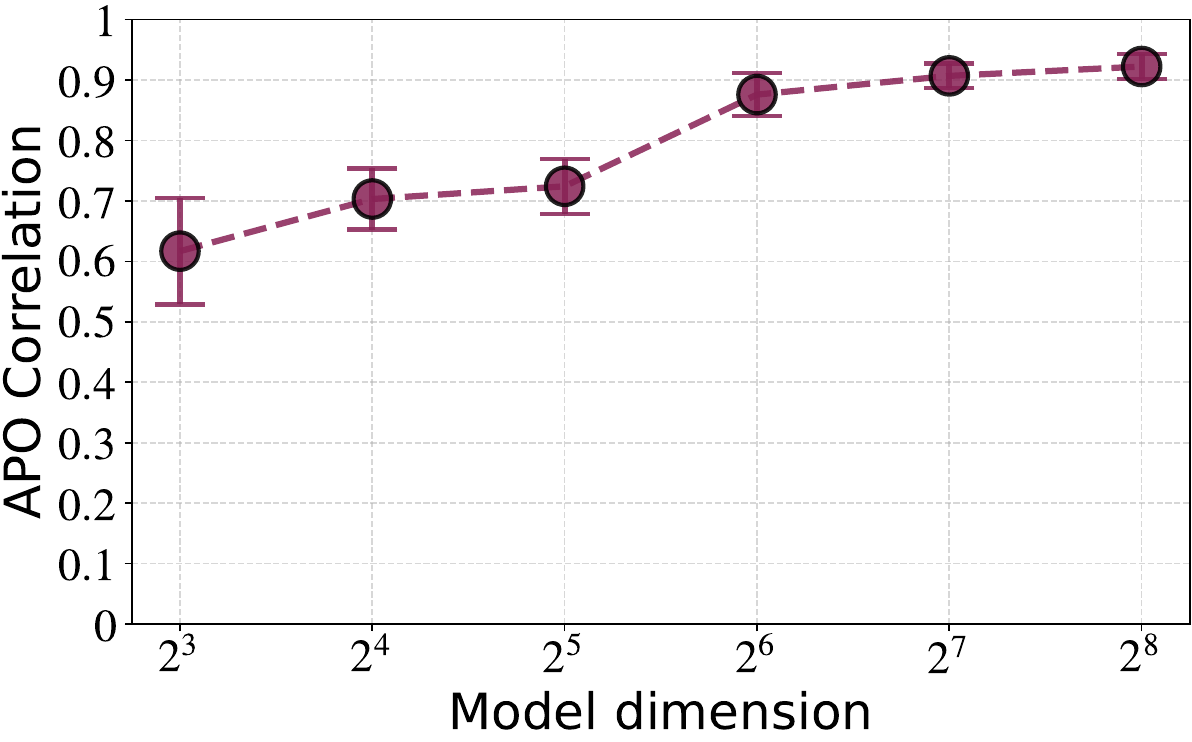}
  \end{subfigure}
  \hfill
  \begin{subfigure}[t]{.32\textwidth}
    \centering
    \includegraphics[width=\linewidth]{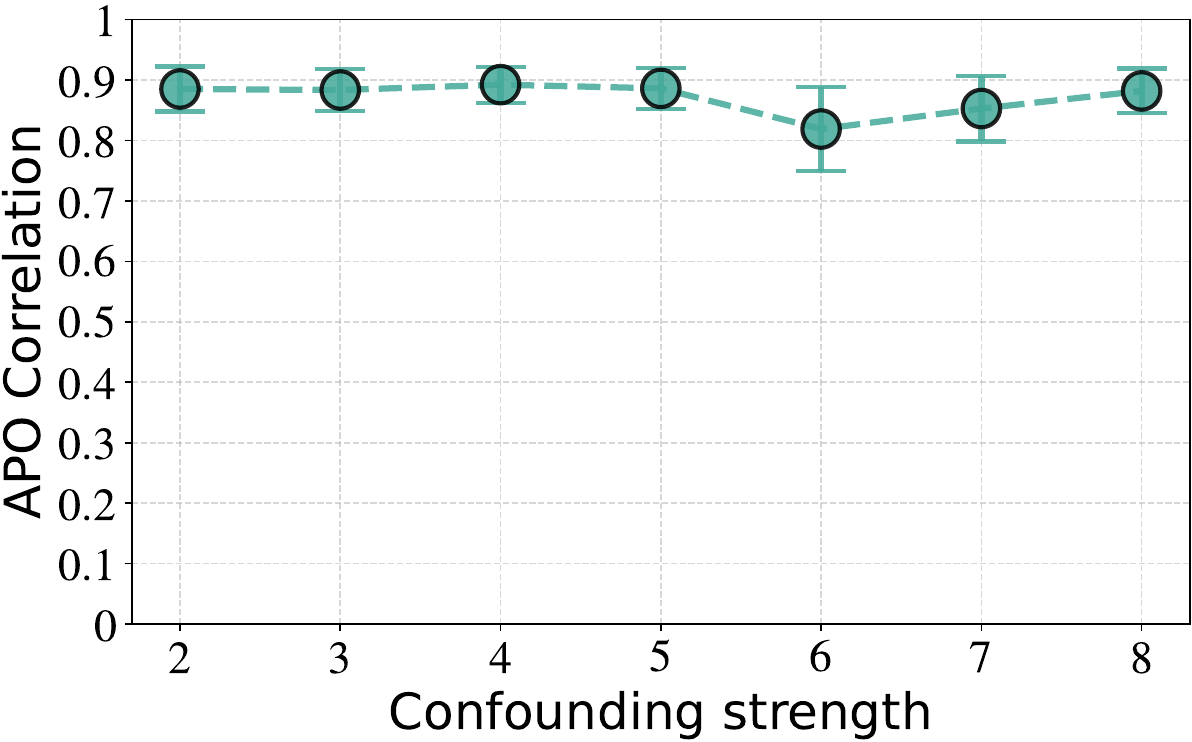}
  \end{subfigure}
\caption{Consistent with standard risk minimization from samples, SW-CRM performance improved with sample size (\textbf{Left}) and model size (\textbf{Middle}) for the synthetic discrete dataset with transformer models. Given observed confounders, performance remained consistent with the strength (or magnitude) of confounding (\textbf{Right}), with increase in standard error at higher strengths.}
\label{fig:ablate_synthetic}
\end{figure}

\begin{figure}[t]
    \centering
    \includegraphics[width=0.75\linewidth]{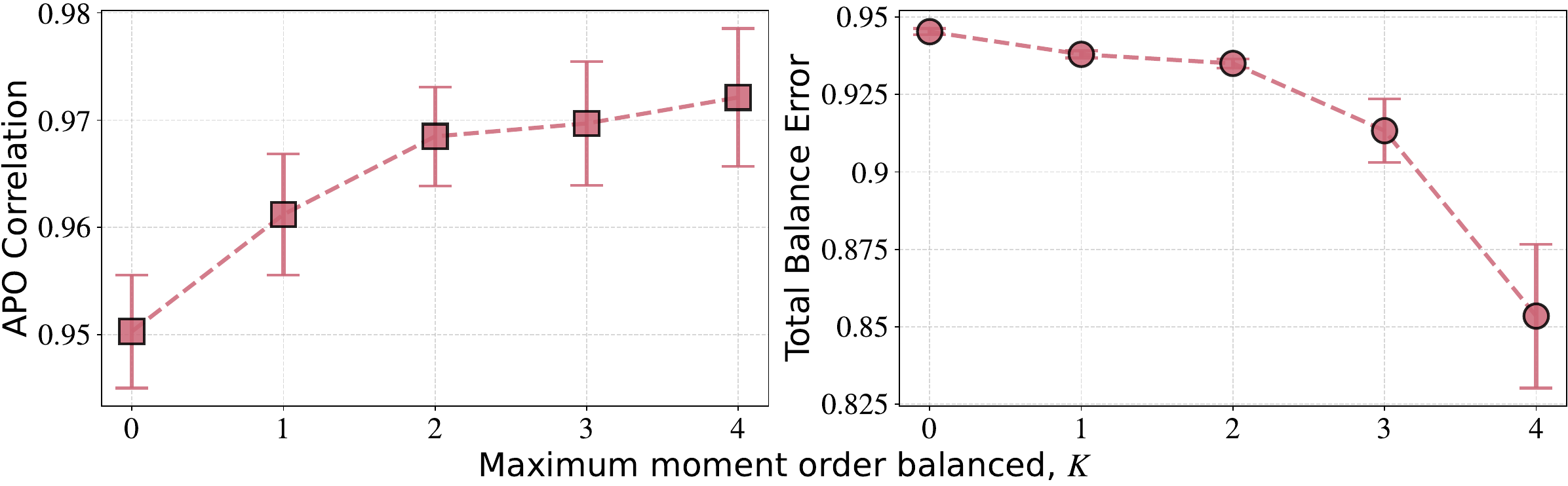}
    \caption{With discrete covariates, balancing higher-order moments directly reduced balance error and hence, APO estimation error, thereby improving correlation with ground-truth.}
    \label{fig:moments_synthetic}
\end{figure}

For the best-performing method, SW-CRM, we conducted further ablation studies to tease apart the effects of different experimental conditions on the correlation between predicted and true APOs, which are visualized in \Cref{fig:ablate_synthetic}. Performance steadily improved with sample size as well as model size, yielding better correlation and lower standard errors. To systematically increase model size, we increased the transformer embedding dimension and its feedforward hidden dimension in proportion. Performance remained stable with increasing confounding strength, suggesting that the estimator was able to correct for large amounts of confounding as long as all confounders were known and the assumption of no unmeasured confounding was satisfied. Finally, we studied the impact of the maximum moment order balanced, $K$, on APO correlation and the total balance error. \Cref{fig:moments_synthetic} shows that as moments of higher orders are included in the balance-error regularizer, APO correlation monotonically increased and correspondingly, the total sum of balance errors up to order-$4$ decreased. This is consistent with our APO error decomposition for non-binary discrete covariates.


\begin{table}[t]
    \centering
    \scriptsize
    \caption{Projecting a single high-dimensional APO estimator for Amazon Reviews onto lower-dimensional treatment attributes (rating, sentiment, length) achieves near-perfect correlation with ground-truth, outcompeting separate estimators that were retrained per attribute.}
\resizebox{\linewidth}{!}{
\begin{tabular}{llcccccc}
\toprule
\multicolumn{2}{c}{} & \multicolumn{4}{c}{\textbf{Varying true APOs}} & \multicolumn{2}{c}{\textbf{Near-constant true APOs}} \\
\cmidrule(lr){3-6} \cmidrule(lr){7-8}
\multicolumn{2}{c}{} & \multicolumn{2}{c}{\textbf{Rating: $|\mathcal{T}| = 5$}} & \multicolumn{2}{c}{\textbf{Sentiment: $|\mathcal{T}| = 2$}} & \multicolumn{2}{c}{\textbf{Length: $|\mathcal{T}| = 2$}} \\
\cmidrule(lr){3-4} \cmidrule(lr){5-6} \cmidrule(lr){7-8}
\multicolumn{2}{l}{\textbf{Estimator}} & \textbf{Rel.\ MAE} & \textbf{Correlation} & \textbf{Rel.\ MAE} & \textbf{Correlation} & \textbf{Rel.\ MAE} & \textbf{Correlation} \\
\midrule
\multirow{2}{*}{OI (Oracle)} & Projected & 0.118 & 0.998 & 0.033 & \textbf{1.000} & 0.091 & -1.000 \\
 & Retrained & 0.359 & 0.983 & 0.012 & 1.000 & 0.010 & 1.000 \\
\midrule
\multirow{2}{*}{OI-CRM} & Projected & 0.103 & \textbf{0.999} & 0.038 & \textbf{1.000} & 0.099 & -1.000 \\
 & Retrained & 0.245 & 0.996 & 0.021 & 1.000 & 0.040 & 1.000 \\
\midrule
\multirow{2}{*}{IPW-CRM (K=1)} & Projected & 0.778 & 0.976 & 0.897 & \textbf{1.000} & 0.903 & \textbf{1.000} \\
 & Retrained & 0.767 & 0.977 & 0.510 & 1.000 & 0.899 & 1.000 \\
\midrule
\multirow{2}{*}{SW-CRM (K=1)} & Projected & 0.129 & 0.991 & 0.119 & \textbf{1.000} & 0.091 & \textbf{1.000} \\
 & Retrained & 0.091 & 0.998 & 0.008 & 1.000 & 0.075 & 1.000 \\
 \midrule
\multirow{2}{*}{SW-CRM (K=2)} & Projected & \textbf{0.057}  & \textbf{0.999} & \textbf{0.007}  & \textbf{1.000}  & \textbf{0.089} & \textbf{1.000}  \\
 & Retrained & 0.042  & 0.999 & 0.006 & 1.000 & 0.075 & 1.000 \\
\bottomrule
\end{tabular}
}
    \label{tab:projections}
\end{table}

\subsection{Do our methods scale to text treatments and large language models?}
\label{sec:text}
To evaluate our methods with treatments in the very high-dimensional and discrete space of all strings, we used the Amazon Reviews dataset, treating each individual review as a treatment, and finetuned Gemma-3-270M to train different components of our estimators. Note that in this more realistic setting, each treatment is observed in only one datapoint and APO estimation relies on the language model's ability to generalize to held-out treatments, hence empirically testing \cref{assump:realizability}. 
\Cref{tab:combined} shows similar trends in estimators' performance to those of the synthetic dataset. IPW-CRM failed more dramatically in this setting, which is explained by extreme propensity scores in the large treatment space of texts and resulting training instability. 
SW-CRM avoided this instability and performed competitively with OI and OI-CRM even at $K=1$, with higher-order moments ($K=2$) further improving performance.
The confounder here has finite support, ($X \in {0,\dots, 7}$), giving an exact error decomposition with $K=7$. Nevertheless, SW-CRM already performs strongly at $K=2$, suggesting that the practically relevant balancing order may be governed by the effective smoothness of the conditional outcome in $X$, rather than by the maximum polynomial degree required for an exact finite-support representation.
Further, SW-CRM and OI-CRM achieved high correlations with the true APOs without requiring access to confounder values at test time, which is necessary for inference with the standard OI oracle.

\subsection{Can we repurpose a single high-dimensional APO estimator to answer several lower-dimensional causal questions efficiently?}
\label{sec:projections}
Finally, we explored how a single high-dimensional APO estimator may be operationalized and used in practice. A straightforward use-case is to compare estimated APOs of two or more texts to predict which one would lead to the best outcome of interest. It is also possible to ask lower dimensional causal questions about specific attributes of the text and answer them by projecting the high-dimensional APO estimate down from the space of texts to the space of the attribute in question, using \cref{eq:proj}. For instance, in the case of Amazon Reviews, we maybe interested in the effects of the review rating ($1, \dots, 5$), sentiment (positive or negative), or length (greater or less than 100 words) on the outcome of purchase. For each estimator, we computed the projected APOs for these three treatment spaces and compared them against their separately retrained attribute-specific counterparts. 

\Cref{tab:projections} shows that the projected estimates from a single high-dimensional APO estimator have competitive or better performance than their retrained counterparts, with SW-CRM ($K=2$) outperforming other variants. This performance is consistent across different estimators for the Rating and Sentiment treatments, with no additional training cost for each attribute. 
The case of Length provides a challenging task with two possible treatments (\emph{i.e.} possible correlations of $-1$ or $1$) with near-constant true APOs. This means that small errors in predictions can reverse the treatment ordering (\emph{i.e.} $-1$ correlation), which was the case with projected estimates of OI and OI-CRM. 
Overall, these results show that a single high-dimensional APO estimator can serve as an efficient tool for answering multiple downstream causal queries simultaneously, without retraining separate estimators for each, as long as the assumption of no unmeasured confounding is satisfied and the lower-dimensional attributes satisfy \cref{assump:projection}.
\section{Limitations and Conclusions}
\label{conclusions}
In this work, we recast average potential outcome (APO) estimation with high-dimensional treatments as causal risk minimization (CRM) over the input space of treatments, overcoming challenges of generalization to unseen treatments. We presented an APO error decomposition into moment-balancing errors to learn accurate targets for risk minimization and showed how to efficiently answer multiple causal questions about treatment attributes by projecting a single high-dimensional estimator onto lower-dimensional attributes, with no additional training cost.

Our work has several limitations that may impact its application, including those shared with any observational study:
\begin{enumerate*}[label=(\roman*)]
    \item the strong ignorability assumption is necessary, though untestable in practical scenarios;
    \item optimization of higher-order moment-balancing errors translates to improved performance but also incurs additional computational costs that is linear in $K$;
    \item generalization of CRM estimators to held-out treatments relies on the realizability of $g$ in the chosen model class, which is not fully testable and is subject to the usual pitfalls of out-of-distribution queries; and
    \item our lower-dimensional projected APOs require access to the relationship between the high-dimensional treatment and its attributes, $p_{T'|T}$, along with the conditional independence assumption that requires $T'$ to depend only on $T$.
\end{enumerate*}

When the required assumptions are satisfied, our results empirically verify the analytic APO error decomposition and demonstrate performance trends with scaled dataset and model sizes that are consistent with standard risk minimization. The semi-synthetic Amazon Reviews experiments show successful application of CRM estimators to real text treatments and \emph{post hoc} lower-dimensional projections that are competitive with attribute-specific retraining, unlocking the potential for real-world causal estimation in domains with truly high-dimensional treatments.

\subsubsection*{Broader Impact Statement}
In addition to the methodological limitations discussed above, this work has applications with societal impacts that deserve thoughtful consideration before any real-world deployment. It is important to establish the reliability and robustness of estimates before using them to make decisions, especially in high-stakes domains. In applications like healthcare, where the training data required for the CRM estimators usually include personal and/or health information, collecting, storing, and training models on this data requires accounting for privacy concerns. There is also potential to use text treatment effect estimators like the ones proposed here in manipulative or deceptive use-cases, for instance generating or selecting fake product reviews that maximize purchase likelihood, framing financial disclosures to mislead investors, or optimizing political or public health messaging to exploit the vulnerabilities of particular audiences. To mitigate these risks, we recommend validation against held-out randomized data where feasible, transparency about when text interventions have been optimized using causal estimators, and adherence to domain-specific ethical and regulatory standards.

\subsubsection*{Acknowledgments}
We would like to thank Leonardo Cotta for feedback on a draft of the paper and Roger Grosse and Rahul Krishnan for helpful discussions.
Resources used in preparing this research were provided in part by the Province of Ontario, the Government of Canada through CIFAR, and companies sponsoring the Vector Institute. We acknowledge the support of the Natural Sciences and Engineering Research Council of Canada (NSERC), RGPIN-2021-03445.

\bibliography{main}
\bibliographystyle{tmlr}

\newpage
\appendix
\crefalias{section}{appendix}
\section*{Appendix}
\section{Notation}
\label{app:notation}
\bgroup
\def\arraystretch{1.5}
\begin{tabular}{|p{0.5in}|p{4.5in}|}
\hline
$\displaystyle X$ & Random variable corresponding to confounders.\\
$\displaystyle T$ & Random variable corresponding to high-dimensional treatment.\\
$\displaystyle T'$ & Random variable corresponding to low-dimensional treatment attribute.\\
$\displaystyle Y$ & Random variable corresponding to outcome observed .\\
$\displaystyle x$ & Possible instance of $X$ from its support $\mathcal{X}$.\\
$\displaystyle t$ & Possible instance of $T$ from its support $\mathcal{T}$.\\
$\displaystyle t'$ & Possible instance of $T'$ from its support.\\
$\displaystyle y$ & Possible instance of $Y$ from its support $\mathcal{Y}= \{0, 1\}$ or $\mathbb{R}$.\\
$\displaystyle Y(t)$ & Random variable corresponding to potential outcome under treatment $t$.\\
$\displaystyle X_i$ & Sampled value of $X$ for individual $i$.\\
$\displaystyle T_i$ & Sampled value of $T$ for individual $i$.\\
$\displaystyle T'_i$ & Sampled value of $T'$ for individual $i$.\\
$\displaystyle Y_i$ & Sampled value of $Y$ for individual $i$.\\
$\displaystyle g(t)$ & Average potential outcome (APO) given by $\mathbb{E}[Y(t)]$, where the expectation is over some defined population of individuals.\\
$\displaystyle g'(t')$ & Lower-dimensional APO under treatment $t'$.\\
$\displaystyle \hat{g}(\cdot)$ & Learned APO estimator.\\
$\displaystyle n$ & Total number of data points.\\
$\displaystyle \hat{e}(t, x)$ & Propensity score estimator, to approximate $p_{T \mid X}(t \mid x)$.\\
$\displaystyle \hat{f}(t, x)$ & Conditional outcome estimator given treatment and confounder, to approximate $\mathbb{E}[Y | T=t, X=x]$.\\
$\displaystyle \hat{w}(t, x)$ & Stabilized weights estimator, to approximate $\frac{p_T(t)}{p_{T \mid X}(t \mid x)}$.\\
$\displaystyle \varepsilon_k(t)$ & Order-$k$ balancing error.\\
$\displaystyle K$ & Maximum moment order included in balancing error optimization.\\

\hline
\end{tabular}
\egroup
\newpage
\section{Derivations}
\label{app:derivations}

\subsection{APO error decompositions into balancing errors}
\label{app:error_decomp}
Recall that under \cref{assump:poly}, we can write $f(t,X) = \sum_{k=0}^K c_k(t)\,X^k)$ for constants $c_k(t)$. 
Then, using \cref{eq:f_balance}, we have:
\begin{equation}
  \hat{g}(t) - g(t) = \sum_k c_k(t) \cdot \varepsilon_k(t), \text{ where } 
  \varepsilon_k(t) = \mathbb{E} \left[\hat{w}(T,X) \cdot X^k \mid T=t\right] - \mathbb{E}[X^k],
  \label{appeq:general-decomp}
\end{equation}
is the \emph{order-$k$ balancing error} corresponding to $X^k$. We now instantiate this decomposition for confounders in different settings: binary, discrete, and continuous.

\paragraph{Binary X (warm-up).}
If $X \in \{0,1\}$, we have $f(t,X) = f(t,0) \cdot 1 + (f(t,1) - f(t,0)) \cdot X$. Substituting into \cref{appeq:general-decomp}, the APO error depends on two balance error terms:
\begin{equation}
  \hat{g}(t) - g(t) = f(t,0) \cdot \varepsilon_0(t) + \big[f(t,1) - f(t,0)\big] \cdot \varepsilon_1(t),
  \label{eq:binary-decomp}
\end{equation}
where $\varepsilon_0(t) = \mathbb{E}[\hat{w} \mid T=t] - 1$ is a normalization error and $\varepsilon_1(t) = \mathbb{E}[\hat{w}(t, X)\cdot X \mid T=t] - \mathbb{E}[X]$ is the common first moment balance error. In this case, the APO error of the risk minimizer vanishes whenever the weights are properly normalized and balance the first moment of $X$.

\paragraph{Discrete finite $X$.}
For $X \in \{0, 1, \ldots, K\}$, the Newton forward difference formula \citep{graham1994concrete}, $f(t, \cdot)$ admits the finite expansion,
\begin{equation}
  f(t, X) = \sum_{k=0}^{K} \frac{\Delta^k f(t,0)}{k!} \prod_{j=0}^{k-1} (X-j),
\end{equation}
where $\Delta^k f(t,0) = \sum_{j=0}^k (-1)^{k-j}\binom{k}{j}f(t,j)$ is the $k$-th forward difference at zero and the product $\prod_{j=0}^{k-1} (X-j)$ is also known as the falling factorial. 
Notice that we may collect terms to rewrite this expansion in terms of powers of $X$:
$f(t, X) = \sum_{k=0}^{K} c_k(t) X^k$ for some $c_k(t)$.
Hence,
\begin{equation}
  \hat{g}(t) - g(t) = \sum_{k=0}^{K} c_k(t) \cdot (\mathbb{E} [\hat{w}(t, X) \cdot X^k \mid T=t ] - \mathbb{E} [X^k ]).
\end{equation}
In other words, balancing the moments $\mathbb{E}[X^k]$ for $k = 0, \dots, K$ drives the APO error to zero. 

\paragraph{Continuous $X$.}
Finally, when $X \in \mathbb{R}$ and $f$ is a continuous polynomial of order at most $K$, the Taylor expansion of the true conditional outcome function $f(t,x) = \mathbb{E}[Y \mid T=t, X]$ around $X = 0$ yields for each $t$:
\begin{equation}
  f(t,X) = \sum_{k=0}^{K}\frac{f^{(k)}(t,0)}{k!}\,X^k \implies g(t) = \mathbb{E}[f(t, X)] = \sum_{k=0}^K\frac{f^{(k)}(t,0)}{k!}\,\mathbb{E}[X^k],
\end{equation}
where $f^{(k)}(t,0) = \frac{\partial^k f(t,x)}{\partial x^k}\big|_{x=0}$. 
Similarly, the estimated APO is:
\begin{equation}
  \hat{g}(t) = \mathbb{E}[\hat{w}(T,X) \cdot f(t, X) \mid T = t] = \sum_{k=0}^{\infty}\frac{f^{(k)}(t,0)}{k!}\,\mathbb{E}\!\left[\hat{w}(T,X)\cdot X^k \mid T=t\right].
\end{equation}
Let the order-$k$ balancing error be $\varepsilon_k(t) = \mathbb{E}[\hat{w}(T,X)\cdot X^k \mid T=t] - \mathbb{E}[X^k]$. Then,
\begin{equation}
  \hat{g}(t) - g(t) = \sum_{k=0}^{\infty}\frac{f^{(k)}(t,0)}{k!}\cdot\varepsilon_k(t).
\end{equation}

Across all cases, the APO estimation error of the risk minimizer can be written in terms of moment balancing errors of the confounders.

\subsection{Deriving lower-dimensional projections of high-dimensional APOs}
\label{app:projection}
Let $T'$ denote the lower-dimensional attribute of the high-dimensional treatment $T$, where the relationship between the two variables is specified through a known conditional distribution $p_{T' \mid T}$, so that $p_{T'}(t') = \bbE_T [p_{T'|T}(t'|T)]$. We assume $T'$ is independent of $X$ and $Y(\cdot)$ given $T$.

To relate the APOs $g(\cdot)$ over the support of $T$ to the APOs $g'(\cdot)$ over the support of $T'$, consider the ideal completely randomized experiment (CRE) where the assignment distribution over $T$ induces an assignment distribution over $T'$. The randomized distribution over all variables is given by
\begin{equation*}
    p_{\text{cre}}(y, x, t, t') =  p_{Y(t), X}(y, x) p_{T}(t) p_{T' | T}(t' | t).
\end{equation*}
Under this distribution, we can estimate joint-treatment APOs under $T=t$ and $T'=t'$ as:
\begin{align*}
    g(t,t') &= \frac{1}{n} \sum_{i=1}^n \frac{Y_i\indicator{T_i = t, T'_i = t'}}{p_{T, T'}(T_i, T'_i)} \\
    &\approx  \sum_{x,s,s'} \int\frac{y \indicator{s=t,s'=t'}}{p_{T,T'}(s,s')}p_{\text{cre}}(y, x, s, s') \, dy \\
    &= \bbE_{X,Y|T=t, T'=t'} [Y | T=t, T'=t'] \\
    &= \bbE_{X,Y|T=t} [Y | T=t] \\
    &= g(t),
\end{align*}
where the second-to-last equality follows from the conditional independence of $T'$ and $(X, Y(\cdot))$ given $T$.
Hence, under the randomized distribution over $T'$, we have APOs given by:
\begin{align*}
    g'(t') &= \bbE_{X,Y|T'=t'} [Y | T'=t'] \\
    &= \bbE_{T| T'=t'} [ \bbE_{X,Y|T'=t', T} [Y | T'=t', T] \\
    &= \bbE_{T| T'=t'} [g(T, t')] \\
    &= \bbE_{T| T'=t'} [g(T)].
\end{align*}
Our experiments estimate $g$ with different APO estimators $\hat{g}$ and the expectation above with the Monte Carlo estimate:
\begin{equation*}
    \hat{g}'(t') = \frac{\sum_{i=1}^n \indicator{T'_i=t'} \, \hat{g}(T_i)}{\sum_{i=1}^n \indicator{T'_i=t'}}.
\end{equation*}

\newpage
\section{Dataset Details}
\label{app:dataset_details}

\subsection{Linear Continuous Setting}
\label{sec:gaussian_details}
We generated data according to:
\begin{equation*}
\begin{aligned}
    X_i &\sim \cN(0, 1), \\
    T_i \mid X &\sim \cN(X, 1), \\
    Y_i &= 1 + 2 T + 3 X + \epsilon,
\end{aligned}
\end{equation*}
where $\epsilon \sim \cN(0, 1)$ is independent noise. The true APOs are given by $g(t) = 1 + 2 t + 3 \, \bbE[X]$.

The APO error decomposition into the order-$0$ and order-$1$ moment-balancing error follows from the continuous case analyzed in \cref{error_decomp}, or can be directly derived for this data generating process.
\begin{equation*}
\begin{aligned}
    \hat{g}_{\text{SW-CRM}}(t) - g(t)
    &= \bbE[\hat{w}(T, X) Y \mid T=t] - g(t) \\
    &= \bbE[\hat{w}(T, X) \mid T{=}t] + 2 t \bbE[\hat{w}(T, X) \mid T=t] \\
    &\quad + 3 \bbE[\hat{w}(T, X) X \mid T=t] + \bbE[\hat{w}(T, X) \epsilon \mid T=t] \\
    &\quad - (1 + 2t + 3 \, \bbE[X]) \\
    &= (1+2t) (\bbE[\hat{w}(T, X) \mid T{=}t] - 1) + 3 (\bbE[\hat{w}(T, X) X \mid T=t] - \bbE[X]) \\
    &\quad + \bbE[\hat{w}(T, X) \mid T=t] \, \bbE[\epsilon] \\
    &= (1 + 2 t) \varepsilon_0(t) + 3 \varepsilon_1(t).
\end{aligned}
\end{equation*}

\subsection{Synthetic Discrete Dataset}
\label{sec:discrete_details}
We generated discrete sequence data with length-$4$ confounders and length-$3$ treatments according to:
\begin{equation*}
\begin{aligned}
X_i^{(k)} &\sim \mathrm{Uniform}\bigl({0,\ldots,V^X_k{-}1}\bigr), \quad k = 0,1,2,3, \\
T_i^{(j)} \mid X_i &\sim \mathrm{Categorical}\bigl(p_j(\cdot \mid X_i)\bigr), \quad j = 0,1,2, \\
Y_i \mid T_i, X_i &\sim \mathrm{Bernoulli}\bigl(\sigma(\mu(T_i, X_i))\bigr),
\end{aligned}
\end{equation*}
where ``vocabulary'' sizes were $V^X = (5,4,2,3)$, such that $|\mathcal{X}| = 120$, and $V^T = (4,2,2)$, such that $|\mathcal{T}| = 16$. 
For the model size ablation study, we used $V^T = (16,2,2)$, such that $|\mathcal{T}| = 64$, to demonstrate the benefits of larger models.
The true propensity factorizes across the treatment sequence, $p_{T|X}(t \mid x) = \prod_{j=0}^{2} p_j(t^{(j)} \mid x)$, where
\begin{equation*}
p_j(t^{(j)} \mid x) = \frac{\exp(\ell_j(x,t^{(j)}))}{\sum_{t^{(j)}=0}^{V^T_j-1}\exp(\ell_j(x,t^{(j)}))},
\end{equation*}
and the logits are
\begin{equation*}
\begin{aligned}
\ell_0(x,t^{(0)}) &= t^{(0)}\bigl(x^{(0)} + (x^{(0)})^2 + 2(x^{(0)})^3\bigr), \\
\ell_1(x,t^{(1)}) &= -t^{(1)}\bigl(x^{(3)} + (x^{(3)})^2 + 2(x^{(3)})^3\bigr), \\
\ell_2(x,t^{(2)}) &= t^{(2)}\bigl(x^{(1)} + (x^{(1)})^2 + 2(x^{(1)})^3\bigr).
\end{aligned}
\end{equation*}
To increase confounding strength in our ablation studies, we increased the scalar multiplying the interaction terms involving confounders and treatments.
The true conditional outcome model is
\begin{equation*}
\begin{aligned}
\mu(t,x) &= 0.3t^{(0)} + 0.2t^{(1)} + 0.15t^{(2)} \\
&\quad + 0.4x^{(0)} + 0.10x^{(1)} + 0.25x^{(2)} + 0.15x^{(3)} \\
&\quad + t^{(0)} x^{(0)} + t^{(0)}(x^{(0)})^2 + t^{(0)}(x^{(0)})^3.
\end{aligned}
\end{equation*}
The true APOs are computed by exact enumeration:
\begin{equation*}
g(t) = \bbE[Y(t)] = \frac{1}{120}\sum_{x \in \mathcal{X}} \sigma(\mu(t,x)).
\end{equation*}

\subsection{Amazon Reviews Text Dataset}
\label{sec:amazon_details}
We constructed a semi-synthetic dataset from the Amazon Reviews 2023 corpus \citep{hou2024bridging}, Electronics category, with $n = 10,000$ reviews. The treatment $T$ is the full text of a product review, along with the rating given.
The confounder $X \in {0, \dots, 7}$ is the product's total rating count, discretized into 8 bins with thresholds $(10, 50, 200, 1000, 5000, 20000, 100000)$.
The outcome was the binary variable corresponding to whether or not a customer purchases the product after reading a given review.
Since counterfactual outcomes are unobservable, we used GPT-5.1 to generate the true conditional purchase probabilities for the purpose of evaluation. For each review and each rating count bin, we queried GPT-5.1 with the system prompt:
\begin{quote}
\ttfamily
You are a customer shopping for an electronic product online on Amazon. You get to see the number of ratings the product has and one complete user review, with their rating. Based on both these things, what is the probability (0.0 to 1.0) that you will purchase the product? Reason about how the number of ratings and the particular review affect your probability of purchase: \newline
\newline
Larger number of ratings indicate a popular product, making you more likely to purchase it. \newline
The more positive the review sentiment is, the more likely you are to purchase it. \newline
A detailed and informative review makes your purchase probability more aligned with the review sentiment. \newline
A large number of ratings also reinforces the review sentiment and accordingly your purchase probability. \newline 
Respond with only a single float between 0.0 and 1.0, nothing else. 
\end{quote} 
and a user prompt of the form: 
\begin{quote} 
\ttfamily
Number of Ratings: <bin label> \newline
Rating: <rating>/5 \newline
Review: <review text> \newline
\end{quote}
yielding $\mu(t, k) = P(Y{=}1 \mid T{=}t, X{=}k)$ for each $(t, k)$ pair. 
The observed outcome was sampled as $Y_i \sim \mathrm{Bernoulli}(\mu(T_i, X_i))$.
The true APO was computed as:
\begin{equation*}
g(t) = \bbE[Y(t)] = \sum_{k=0}^{7} P(X{=}k) \, \mu(t, k),
\end{equation*}
where $P(X{=}k)$ is the empirical bin frequency. 

For the lower-dimensional projections of text APOs, we annotated each review with the following three attributes:
\begin{enumerate}
\item Rating: the rating out of $5$, e.g.\ \texttt{"4/5"},
\item Sentiment: \texttt{"very positive"} if rating $> 4$, else \texttt{"negative or neutral"},
\item Length: \texttt{"more than 100 words"} or \texttt{"less than 100 words"}.
\end{enumerate}

\newpage
\section{Experimental Details}
\label{app:exp_details}

All hyperparameters were chosen on a validation split for each dataset and model.

\subsection{Linear Continuous Setting}
We used linear networks for the conditional outcome model and the APO model that maps treatments to APO targets, each trained with the Adam optimizer at learning rate $10^{-2}$ for 2000 steps. The propensity model was a $3$-layer multilayer perceptron with hidden dimension 64 and ReLU activations, mapping $X \in \mathbb{R}$ to two scalar outputs $\mu_\theta(X)$ and $\log \sigma_\theta(X)$, which parameterized the Gaussian propensities $T \mid X \sim \mathcal{N}(\mu_\theta(X),\, \sigma_\theta(X)^2)$. It was trained via maximum likelihood, with and without balance-error regularization, using Adam with learning rate $10^{-3}$ for 2000 steps. 

\subsection{Synthetic Discrete Dataset}
All trained models (propensity, stabilized weights, conditional outcome, and APO) were $2$-layer transformers with embedding dimension $8$, $2$ attention heads, feedforward dimension $32$, trained with AdamW and learning rate $10^{-4}$ on $n = 10,000$ samples. Propensity and stabilized models were trained for 2000 epochs each and APO and outcome models for 3000 epochs with the cross-entropy loss derived in \cref{crm_targets} for binary outcomes.

\subsection{Amazon Reviews Text Dataset}
For all models, we finetuned a pretrained Gemma-3-270M model with AdamW on their corresponding objectives, using the Transformers library \citep{wolf-etal-2020-transformers} on 2 NVIDIA L40 GPUs. Hyperparameters vary per estimator and are summarized in \cref{tab:amazon_hparams}. 

\begin{table}[h]
\centering
\scriptsize
\caption{Hyperparameters for Amazon Reviews experiments.}
\resizebox{0.6\linewidth}{!}{
\begin{tabular}{lcccc}
\toprule
\textbf{Estimator} & \textbf{Model} & \textbf{LR} & \textbf{Epochs} & \textbf{Batch size} \\
\midrule
OI        & Outcome    & $10^{-5}$ & 5  & 16 \\
\midrule
OI-CRM    & Outcome    & $10^{-5}$ & 10 & 16 \\
          & APO        & $10^{-5}$ & 10 & 16 \\
\midrule
IPW-CRM   & Propensity & $10^{-6}$ & 10 & 4  \\
          & APO        & $10^{-6}$ & 10 & 16 \\
\midrule
SW-CRM    & SW         & $10^{-5}$ & 10 & 16 \\
          & APO        & $10^{-5}$ & 5  & 16 \\
\bottomrule
\end{tabular}
}
\label{tab:amazon_hparams}
\end{table}
\end{document}